\documentclass{esannV2}
\usepackage{graphicx}
\usepackage[latin1]{inputenc}
\usepackage{amssymb,amsmath,array}
\usepackage{svg}
\usepackage{wrapfig}
\usepackage{enumitem}
\usepackage[compact]{titlesec}
\titlespacing*{\subsection}{0pt}{*1}{*0}
\usepackage[numbers]{natbib}
\begin{document}
\title{JEPA for RL: Investigating Joint-Embedding Predictive Architectures for Reinforcement Learning}

\author{Tristan Kenneweg$^1$, Philip Kenneweg$^1$ and Barbara Hammer$^1$
%
\thanks{Acknowledgments. The authors were supported by SAIL. 
SAIL is funded by the Ministry of Culture and Science of the State of North Rhine-Westphalia under the grant no NW21-059A.}
%
\vspace{.3cm}\\
%
1- University of Bielefeld - Technical Faculty \\
Universitaetsstrasse 25, 33615 Bielefeld - Germany
%
\vspace{.1cm}
}

\maketitle

\begin{abstract}
Joint-Embedding Predictive Architectures (JEPA) have recently become popular as promising architectures for self-supervised learning. Vision transformers have been trained using JEPA to produce embeddings from images and videos, which have been shown to be highly suitable for downstream tasks like classification and segmentation. In this paper, we show how to adapt the JEPA architecture to reinforcement learning from images. We discuss model collapse, show how to prevent it, and provide exemplary data on the classical Cart Pole task.

\end{abstract}

\section{Introduction}

Reinforcement learning from images is often a slow and compute-intensive process since an image is a very high-dimensional state description \cite{mnih2015human}. The actual information needed from a state is often much lower-dimensional. In the classic Cart Pole task \cite{towers2024gymnasium}, the image state at typical resolution has $d_{img}=400\times600\times3 = 720,000$ dimensions. But the actual state as given by the simulation only consists of cart position, angle, velocity and angular velocity, resulting in $d_s=4$ dimension. 

Consequently, it is desirable to learn a low-dimensional representation from images on which reinforcement learning can take place \cite{curl}. This representation has to capture all necessary information to master a given task. Towards this end, many techniques have been developed; one of the most famous is to use a variational autoencoder \cite{kingma2014auto}.

In an autoencoder setup an image is fed to a network with a bottleneck in the middle that contains a very limited number of $n_a$ neurons. The network is trained with a reconstruction loss between original and predicted pixel values. For images that have a large amount of repetitive structure in them, this approach works exceedingly well \cite{ha2018world}.
However, autoencoders have a key limitation in that they assign equal value to every pixel. In the Cart Pole example, that means that a white background pixel is equally important as a pixel of the pole, although the latter provides more relevant information. This can lead to results in which the relevant moving parts of the images are blurry and difficult to recognize while the background is perfectly crisp \cite{Burgess:2018hqi}. 
\begin{wrapfigure}{r}{0.5\textwidth}
\centering
\includegraphics[width=0.48\textwidth]{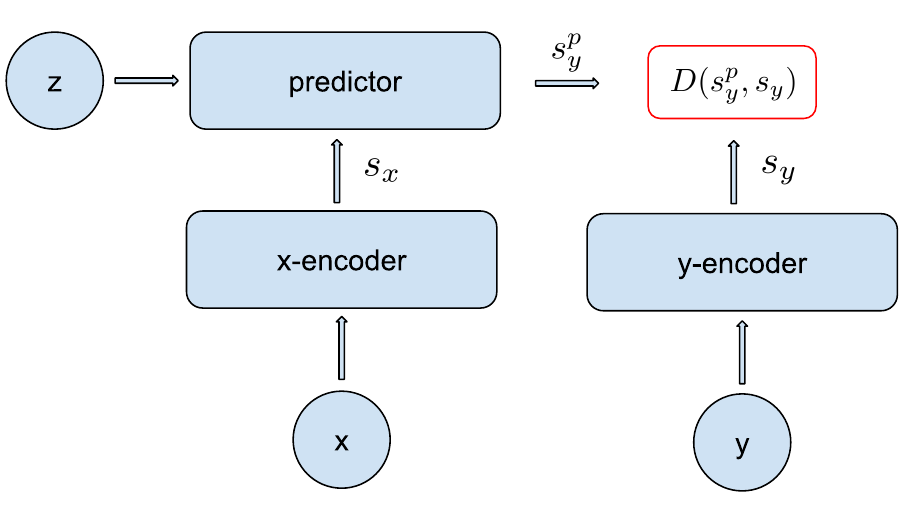}
\caption{Overview of JEPA as proposed by Yann LeCun. }\label{Fig:jepa}
\vspace{-0.02\textwidth}
\end{wrapfigure}
To overcome this limitation, Yann LeCun \cite{lecun2022path} proposed the Joint-Embedding Predictive Architecture (JEPA) as seen in Figure \ref{Fig:jepa}. In JEPA, separate context and target encoder networks encode information that is spatially or temporally close, for example, different patches in an image or different frames in a video. A shallow predictor network predicts the target encoding from the context encoding, given an additional latent variable $z$.  This architecture has the advantage that it is trained entirely by reconstruction error in latent space and can thus choose to create embeddings that ignore irrelevant details in an image. The downside is that it is much more prone to collapse, since a constant output of the context and target encoders, along with a predictor that performs an identity operation, will result in a minimal loss.

In this paper, we show how to adapt JEPA to reinforcement learning problems that can be described with a low-dimensional state. By this, we refer to problems like Atari games, where each state can be described with a single vector that typically has fewer than 100 entries. Our main contributions are: 

\begin{itemize}[noitemsep, topsep=0pt]
    \item We explain in detail how to use a vision transformer in tandem with JEPA to learn embeddings which can be used for successful reinforcement learning.
    \item We discuss possible model collapse scenarios and show how to avoid them. 
    \item We show exemplary data using the classical Cart Pole task. 
\end{itemize}

\section{Methods}

\begin{figure}[h]
\centering
\includegraphics[scale=0.6]{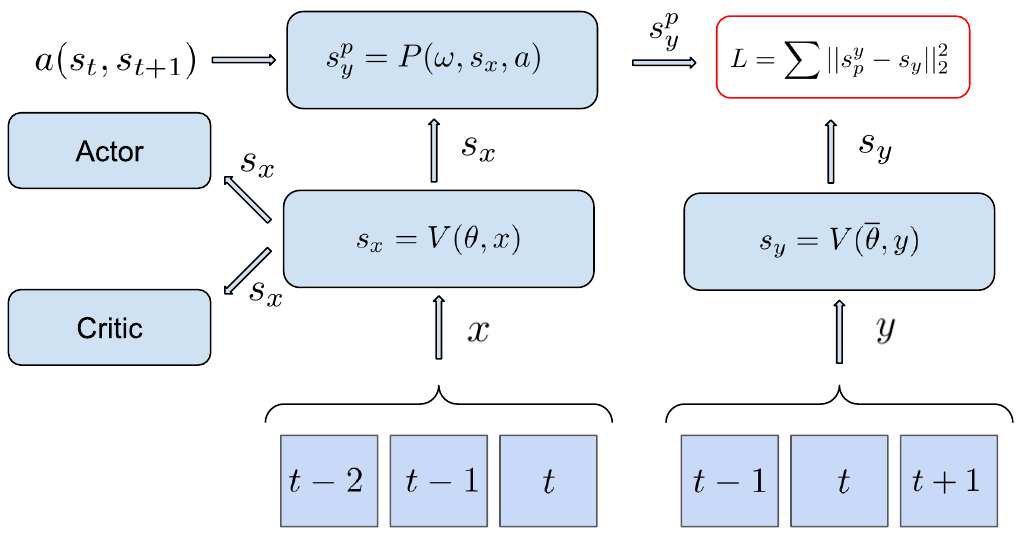}
\caption{Overview of our JEPA pipeline as adapted to reinforcement learning. We feed all patch embedding of frames $f_{t-2}$ to $f_{t}$ into the x-encoder $V(\theta,x)$.}

\label{Fig:rljepa}
\end{figure}

In this Section, we explain in detail how we adapted JEPA to reinforcement learning problems. We focus on reinforcement learning tasks that provide image input and no further state information. Our reinforcement learning networks (usually actor and critic) solely rely on the produced embeddings. See Figure \ref{Fig:rljepa} for an overview of our architecture.

\subsection*{Input and Encoder}
Since our agent relies solely on x-encoder embeddings, these must encapsulate all task-relevant information, including temporal context. Therefore, we encode the last three frames $x = f_{t-2}, f_{t-1}, f_{t}$ of a given simulation. For the x-encoder, we choose a vision transformer $V(\theta,x)$. In each forward pass we feed the patch embeddings of all images to the transformer and add a positional encoding that encodes not only the $i,j$ patch coordinates in the images but also the $t$ coordinate, which indicates from which frame a given patch is taken.

For our target $y$ value, we choose the frames $y = f_{t-1}, f_t, f_{t+1}$. We do this because we want to force the embeddings of $x$ to encode all the information necessary to easily predict the embedding of the next frame. We choose to encode $f_{t-1}, f_t$, and $f_{t+1}$ instead of just $f_{t+1}$ because we use the same architecture for our y-encoder. We set the weights of our y-encoder $\overline{\theta}$ to a running average of the x-encoder weights: $\overline{\theta}_{t+1} = 0.99 \cdot \overline{\theta}_t + 0.01 \cdot \theta_{t+1}$
This approach has been shown to prevent collapse in previous work \cite{grill2020bootstrap}. We initialize both networks with the same values. We do not pass gradient updates through the y-encoder.

\subsection*{Predictor}
We choose a shallow two-layer MLP as our predictor. We keep our predictor intentionally small so that the task of state prediction is solved in the embedding stage and not by the predictor. To have all the necessary information to make a prediction, the predictor needs to be fed the action taken by the actor to get from state $s_t$ to state $s_{t+1}$. We project the one-hot encoded action to the dimension of the hidden layer using a linear layer and add it to the embedding after the first layer.

\subsection*{Learning Objective}
\label{sec:learningobj}
There are many valid choices for the learning objective. A straightforward one, similar versions of which have shown success on downstream tasks in other work \cite{assran2023self}, is to take the last-layer embeddings of all patches from the x-encoder vision transformer and feed those to the predictor. This would result in a very high-dimensional representation $s_x$ and thus require significant computational resources and data. While this objective is potentially very powerful when used on complex scenarios with massive computational resources, it is not ideal when learning Atari games or similar tasks, since the relevant information can be contained in a much smaller representation.

Instead, we choose to prepend the equivalent to a learnable classification token to the x-encoder vision transformer. \textbf{We only feed the last-layer embeddings of this classification token to the predictor.} Thus, the dimension of $s_x$ is only the embedding dimension $d_{\text{emb}}$. For our experiments, we choose $d_{\text{emb}} = 64$. Since we know that the state of the Cart Pole game can be represented using a four-dimensional vector, this is sufficient. The same is true for all learnable tasks for which a short state description can be created.

Our JEPA loss is the Euclidean distance between the predictor output $s_y^p$ and the y-encoder output $s_y$:
\begin{equation}
    L_{\text{JEPA}} = \left\| \mathbf{s}_{y}^p - \mathbf{s}_{y} \right\|_2^2 
    \label{eq:loss}
\end{equation}

\subsection*{Collapse Prevention}
\label{sec:collapse}
     As mentioned in the introduction, JEPA is prone to collapse. A constant output of the x-encoder and y-encoder, along with an identity operation of the predictor, will result in a perfect loss. An indicator of collapse that we can observe is the mean batch-wise variance of the embeddings $s_x$, which drops to values below $10^{-7}$ in a collapse scenario. To counteract this, we have two methods available:

We can propagate the actor and critic losses. Since we feed $s_x$ into the actor and critic network, we can propagate the losses through our x-encoder. Thus, we give the encoder network an incentive to learn informative representations $s_x$. We do not define specific actor and critic losses, since these can be arbitrarily chosen.

Furthermore, we can add a more direct regularization loss to prevent collapse. We do this by encouraging batch-wise variance:
\begin{equation}
L_{\text{reg}} = -\min \left(1, \frac{1}{d_{\text{emb}}} \sum_i^{d_{\text{emb}}} \text{Var}\left( \mathbf{s}_x \right)_i \right)
\label{eq:regloss}
\end{equation}
We clamp this loss to 1 since variance is an unbounded metric. Here, $\mathbf{s}_x$ refers to a tensor that contains a batch dimension, and $d_{\text{emb}}$ is the embedding dimension. Encouraging variance in embeddings has been shown to be effective in preventing collapse in self-supervised learning \cite{vicreg}.

\subsection*{Gradient Propagation}
The x-encoder can be purely trained from the loss described in Equation~\ref{eq:loss}. However, in practice, it is more effective to back-propagate the actor and critic losses through the x-encoder vision transformer. Thus, the total loss is given by 
\begin{equation}
    L = L_{\text{JEPA}} + L_{\text{actor}} + L_{\text{critic}} + L_{\text{reg}}
    \label{eq:totalloss}
\end{equation}
where $L_{\text{reg}}$, as described in Section~\ref{sec:collapse}, can be added if model collapse is a problem. 

\section{Results}

\begin{figure}[h]
\centering
\includegraphics[width=1\textwidth]{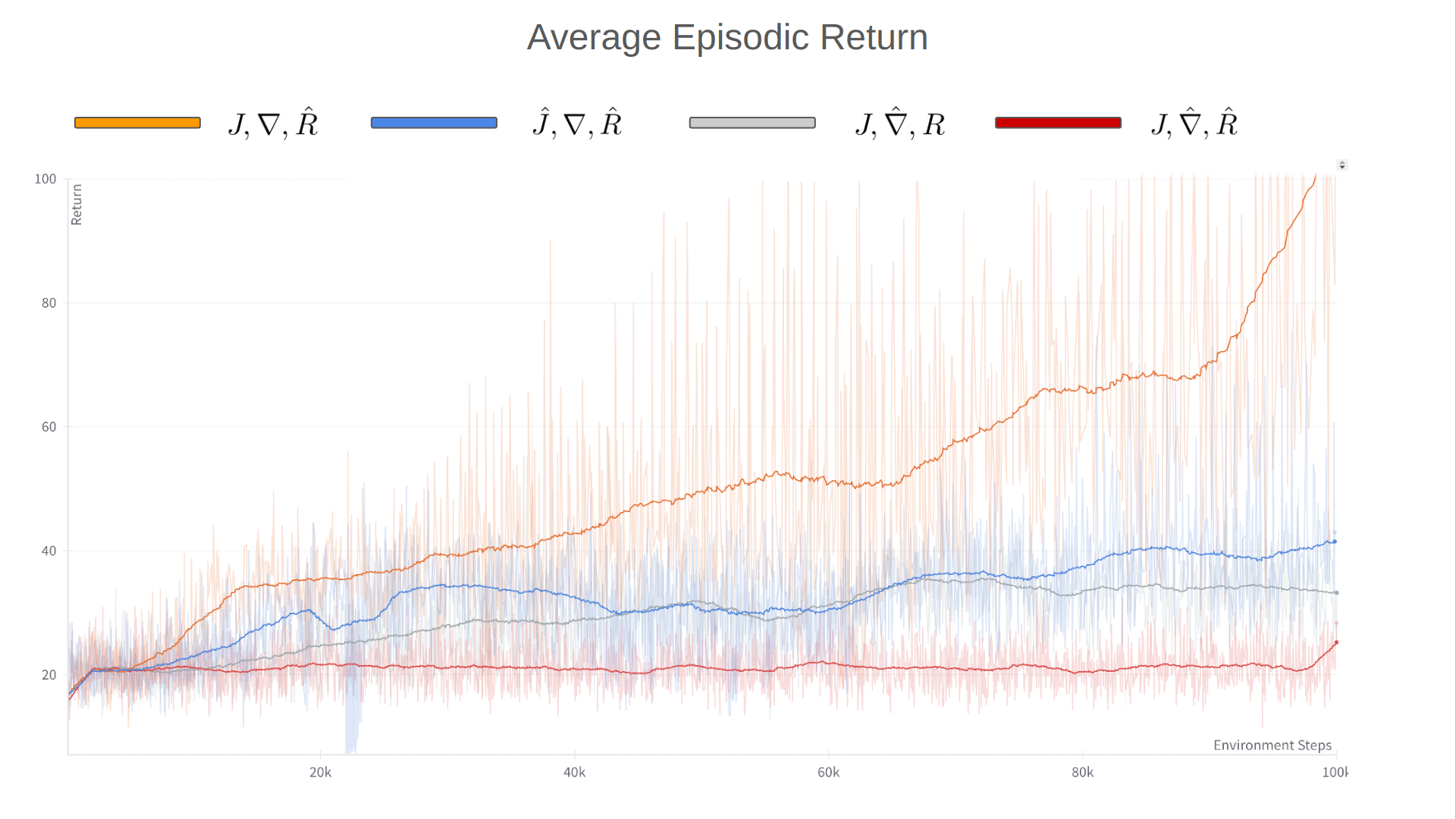}
\caption{Average episodic return over the first 100k environment steps for all four configurations. Each graph shows the accumulated results of 5 runs.}
\label{fig:results}
\end{figure}

We test our framework on the Cart Pole reinforcement learning task using pixel observations and an actor-critic-style PPO algorithm. To evaluate different configurations, we vary three conditions: Including or excluding the JEPA loss (denoted by $J$ or $\hat{J}$), deciding whether reinforcement learning gradients are back propagated to the image encoder (denoted by $\nabla$ or $\hat{\nabla}$), and applying or omitting the regularization loss from Equation \ref{eq:regloss} (denoted by $R$ or $\hat{R}$). We test 4 configurations:

\begin{enumerate}
    \setlength\itemsep{0em} 
    \setlength\topsep{0em} 
    \item $\hat{J},\nabla,\hat{R}$: This is our baseline test in which we omit the JEPA loss and train the encoder purely from the gradients of the actor and critic networks.
    \item $J,\nabla,\hat{R}$: The encoder is trained using both JEPA loss and gradients from the actor and critic.
    \item $J,\hat{\nabla},\hat{R}$: JEPA loss without actor-critic gradient propagation. The actor and critic are still trained using PPO, but we stop the gradient flow when feeding the embeddings to them, so the encoder is trained only via the JEPA loss.
    \item $J,\hat{\nabla},R$: JEPA loss with regularization loss, without actor-critic gradient propagation. We add the regularization loss described in Equation~\ref{eq:regloss} to prevent collapse.
\end{enumerate}

Figure~\ref{fig:results} shows the running average of the episodic return over the first 100k environment steps. Each graph shows the accumulated results of 5 runs. In Cart Pole, one reward is given per frame when the pole is upright.

$\hat{J},\nabla,\hat{R}$: We observe that the agent learns some advantageous behavior, albeit in a limited fashion, even though we omit the JEPA loss. This is reasonable since we still update the x-encoder via the PPO-style actor and critic losses. The embedding variance is in a reasonable range between 0 and 1.

 $J,\nabla,\hat{R}$: When combining JEPA and reinforcement learning losses, we get the best results. The reward increases much faster and does not plateau. We also observe reasonable embedding variances.

$J,\hat{\nabla},\hat{R}$: When stopping the gradients of the actor and critic from back propagating through the JEPA encoder, we observe a model collapse. The encoder maps all inputs to the same embedding, which leads to an ever-decreasing batch-wise variance and a low JEPA loss (not shown here). Since the embeddings do not contain any information, the actor cannot learn anything, and the episodic return never increases. In some cases the embeddings recover from this collapse state when training is continued for longer. 

$J,\hat{\nabla},R$: When the regularization loss, as described in Equation~\ref{eq:regloss}, is added, model collapse is prevented, as shown by the batch-wise variance. The actor can learn from the information contained in the embedding, although much slower as with actor and critic gradients. This shows that JEPA is able to learn informative state representations without gradient propagation from the reinforcement learning task.

We conclude that JEPA can successfully produce embeddings for reinforcement learning from image tasks, especially when the encoder is trained with a combination of JEPA and reinforcement learning gradients.

\section{Conclusion}
In this paper, we presented a method to adapt the Joint-Embedding Predictive Architecture to reinforcement learning from images. We showed how to construct encoder and target-encoder inputs for vision transformers that capture spatio-temporal information using embeddings of appropriate dimensionality. We investigated model collapse and demonstrated how to prevent it using backpropagation of reinforcement learning gradients. Overall, we conclude that JEPAs are promising candidates for reinforcement learning and encourage further work in this direction.


\begin{footnotesize}


\bibliographystyle{unsrt}
\bibliography{references}

\end{footnotesize}


\end{document}